\documentclass[runningheads]{llncs}
\usepackage[T1]{fontenc}
\usepackage{amssymb}
\usepackage{booktabs}
\usepackage{multirow}
\usepackage{amsmath}
\usepackage{cite}
\usepackage{microtype}
\usepackage{float}
\usepackage{placeins}

\usepackage{graphicx}
\begin{document}
\title{Discovering Relationships in Data Lakes Using Large Language Models: An Industrial Case} 
\titlerunning{Discovering Relationships in ERP Data Lakes with LLMs}
%
\author{Ahlame Diouan\inst{1,2}\orcidID{0009-0003-1481-4662} \and
Eric Ferey\inst{2}\and
Sabine Loudcher\inst{1}\orcidID{0000-0002-0494-0169}\and 
Jérôme Darmont\inst{1}\orcidID{0000-0003-1491-384X}}
\authorrunning{A. Diouan et al.}
%
\institute{Université Lumière Lyon 2,
UR ERIC, Lyon, France\\
\email{\{sabine.loudcher,jerome.darmont\}@univ-lyon2.fr}
\and
Bial-X, Lyon, France\\
\email{\{ahlame.diouan,eric.ferey\}@bial-x.com}}
\maketitle      
%
\begin{abstract}
Data lakes rely on metadata to remain usable, yet this metadata is often limited or weakly informative for column relationship discovery, especially in ERP-derived datasets with coded or abbreviated schema labels. We propose ColRel, a two-stage method that builds column embeddings from metadata and data available at ingestion time. In difficult cases, such as coded schemata, business dictionaries help interpret column names and support the generation of short natural-language descriptions used in the second stage. Experiments on public benchmarks and an industrial ERP dataset show that ColRel is particularly effective in semantically related, weak-signal settings.

\end{abstract}
\section{Introduction} 
As organizations integrate data from diverse systems, data lakes have become a common choice to store raw, heterogeneous tables at scale. However, this flexibility shifts the burden to metadata: when documentation is missing or weak, data lakes quickly become difficult to explore, and discovering relationships between tables becomes largely manual. The problem is particularly acute for enterprise resource planning (ERP) data, where information is fragmented across many tables and column names are often abbreviated or coded, making semantics hard to infer from names and small samples alone.

We address this challenge in an industrial public-housing setting, where analysts must inspect tables to identify joinable or semantically related columns, both within a single ERP and across multiple ERP systems with diverging conventions. Our goal is to uncover meaningful column-level relationships to support scalable data lake exploration and integration.

To this end, we introduce \textbf{ColRel}, a two-stage method for column relationship discovery. Stage~1 embeds each column from ingestion-time evidence (schema cues and representative values). When signals are weak, Stage~2 optionally generates concise, dictionary-grounded natural-language descriptions with a large language model (LLM) and injects them into the embedding input. Columns are encoded using a sentence embedding model~\cite{reimers2019sentence}, similarities are computed with cosine similarity, and the top-ranked candidates are returned for validation.

This work makes the following contributions.
\begin{itemize}
\item We propose \textbf{ColRel}, a two-stage method that combines embedding-based retrieval with optional LLM-based semantic enrichment for coded, weak-metadata schemata.
\item We evaluate ColRel against established schema-matching baselines~\cite{madhavan2001generic,do2002coma,melnik2002similarity,zhang2011automatic} using the Valentine framework and additional datasets~\cite{koutras2021valentine}.
\item We apply ColRel to an industrial public-housing ERP dataset and release a synthetic counterpart that preserves its weak-metadata characteristics for reproducible evaluation.
\end{itemize}

The remainder of the paper is organized as follows. Section~\ref{sec:relatedwork} reviews existing work on schema matching, and recent embedding and LLM-based approaches. Section~\ref{sec:problem&approach} presents the problem and ColRel. Section~\ref{sec:experimentsetup} describes the experimental setup. Section~\ref{sec:results} reports results. Section~\ref{sec:conclusion} concludes the paper and outlines directions for future work.

\section{Related Work} 
\label{sec:relatedwork}
Schema matching aims to identify schema elements that refer to the same concept across datasets. It is a central problem in data integration and has been widely studied in the literature~\cite{rahm2001survey,doan2005semantic}. Classical approaches combine schema and instance-level information. In this context, Valentine~\cite{koutras2021valentine} provides reference implementations of several matchers and evaluates them in a dataset-discovery setting where methods return ranked candidate correspondences. Among schema-based techniques, Cupid~\cite{madhavan2001generic} combines linguistic and structural similarity, Similarity Flooding~\cite{melnik2002similarity} propagates similarity across schema graphs and COMA~\cite{do2002coma,engmann2007instance} aggregates multiple matching strategies. Instance-based methods complement these approaches by exploiting column values, for example through value distribution comparison~\cite{zhang2011automatic} or overlap-based measures such as Jaccard–Levenshtein. While effective when column names or values provide clear signals, these methods degrade in data lakes where metadata are missing and column names are often coded.

This challenge is amplified in modern data lakes, where datasets are ingested with minimal documentation. The problem is therefore often framed as dataset or table discovery, where systems aim to retrieve related tables and suggest possible joins. Aurum supports exploration of enterprise data repositories through abstractions of dataset relationships~\cite{fernandez2018aurum}, while benchmarks such as LakeBench evaluate methods for joinability and unionability discovery~\cite{deng2024lakebench}. These works emphasize scalable discovery but also highlight the difficulty of identifying relationships under weak metadata.

To overcome the limitations of purely lexical matching, representation learning approaches encode schema elements into dense vectors and retrieve candidates through nearest-neighbor search. EmbDI learns embeddings for relational integration tasks from relational structures~\cite{cappuzzo2021embdi}. More recently, retrieval frameworks based on pre-trained language models (PLMs) have been proposed for join discovery in data lakes. DeepJoin formulates joinable table discovery, including semantic joins, as an embedding retrieval task using pre-trained language models and column evidence~\cite{dong2022deepjoin}. These approaches scale well and naturally produce ranked candidates, but they still rely on informative column names or value signals. In enterprise settings such as ERP systems, column identifiers are often abbreviated or encoded, and values may consist of short categorical codes, which limits the semantic information captured by embeddings derived from names and small samples.

Recent work explores the use of LLMs to capture richer semantic relationships between schema elements. Some approaches use LLMs directly as matchers by prompting them with schema context. Parciak et al.~\cite{parciak2024schema} study the use of off-the-shelf LLMs for schema matching, while ReMatch~\cite{sheetrit2024rematch} incorporates retrieval to provide additional evidence for LLM reasoning. Other works enrich schema elements before matching. For instance, SMUTF~\cite{zhang2025smutf} generates semantic tags for columns and combines them with lexical and instance-based features in a hybrid matching framework. However, many LLM-based methods invoke the model for each candidate pair of columns, which can become costly when exploring large repositories. Hybrid approaches attempt to balance scalability and reasoning by combining retrieval with LLM reranking. Magneto first retrieves candidate correspondences using a smaller model and then refines them through LLM-based reasoning~\cite{liu2024magneto}.

In contrast, ColRel focuses on column relationship discovery in data lakes rather than strict schema alignment. It follows an embedding-based retrieval paradigm similar to representation-based discovery approaches~\cite{cappuzzo2021embdi,dong2022deepjoin}, but targets weak-metadata ERP environments where lexical signals are limited. ColRel introduces an optional semantic enrichment stage that generates concise, dictionary-grounded descriptions for columns and injects them into the embedding input. Unlike LLM-based matching approaches that invoke the model for each column pair~\cite{parciak2024schema,sheetrit2024rematch} or reranking pipelines such as Magneto~\cite{liu2024magneto}, ColRel uses the LLM only once per column to generate a semantic description. The resulting representation is then embedded and reused during similarity retrieval, preserving scalability while strengthening semantic signals in semantically related scenarios where lexical overlap and value equality are insufficient.

\section{ColRel: Column Relationship Discovery}
\label{sec:problem&approach}
\subsection{Problem Statement}
In ERP-sourced data lakes, identifying meaningful relationships between columns is challenging because schemas are heterogeneous, attribute names are often coded or abbreviated, and metadata are weak. Our objective is to automatically uncover column-level relationships that help users detect whether two tables share related entities, even when lexical or structural signals are limited.

More formally, let $T_1$ and $T_2$ be two tables with column sets $A$ and $B$. We study \textit{column-level relationship discovery}: given only the information typically available at data lake ingestion time---namely column names, data types, values, and intra-metadata---our goal is to identify related column pairs $(A_i, B_j)$ and rank them by confidence.

We focus on two types of relationships that are particularly relevant for ERP-derived data lakes. Two tables are \textbf{joinable} if there exists at least one column pair $(A_i, B_j)$ that can serve as an equality join key, i.e., the columns represent the same concept and their value sets overlap sufficiently for an equality join. For instance, the ERP-coded columns \texttt{ggcdoga} and \texttt{hpcdoga}, as well as \texttt{ggdept} and \texttt{hpdept}, are joinable because they denote the same business attributes and share overlapping values despite opaque naming conventions.

Two tables are \textbf{semantically related} if there exists at least one semantically equivalent column pair, but an equality join fails because values are not verbatim equal, for example due to formatting differences, noise, or heterogeneous conventions. In this case, relationships must be inferred beyond direct value overlap. For example, \texttt{ggdtconstruct} in table \texttt{UGP} and \texttt{date\_construction} in table \texttt{PATRIMOINE\_GIM} refer to the same business concept, namely the construction date of an asset, but their names and value formats differ: the former uses an opaque ERP-coded label and compact date-like values such as \texttt{20022506}, whereas the latter uses an explicit label and a standard date format such as \texttt{1996-06-17}.

This example also illustrates the role of semantic enrichment in ColRel. In Stage~1, \texttt{ggdtconstruct} is represented only through its name, type, and representative values, which provide weak semantic evidence. In Stage~2, dictionary entries such as \texttt{dt = date} and \texttt{construct = construction} guide the generation of a short description such as ``construction date of the asset'', which is appended before embedding and brings the representation closer to \texttt{date\_construction}.

These two relationship types directly reflect the challenges of ERP-derived data lakes. In many cases, join keys are fragmented across tables or hidden behind coded identifiers that require semantic interpretation. When direct value overlap is insufficient, semantic reasoning becomes necessary to reveal the underlying correspondences.

\begin{figure}[t]
    \centering
    \vspace{-0.6em}
    \includegraphics[width=0.86\textwidth]{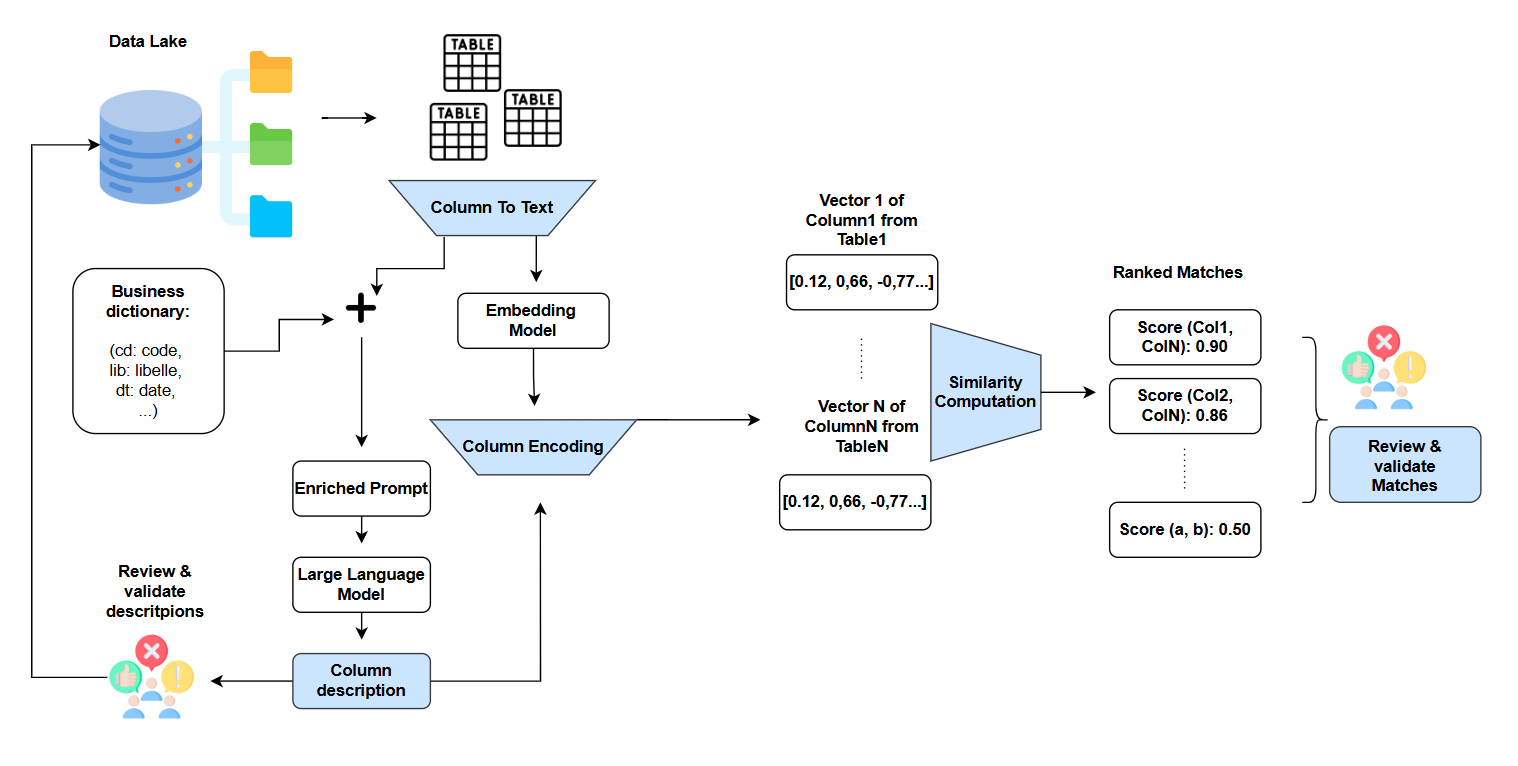}
    \vspace{-0.8em}
    \caption{ColRel two-stage pipeline.}
    \label{fig:pipeline}
    \vspace{-1em}
\end{figure}

\subsection{ColRel Methodology}
\label{sec:ourapproach}
ColRel is a two-stage method for column relationship discovery in data lakes (Figure~\ref{fig:pipeline}). It outputs ranked candidate correspondences that can be inspected and validated by users. The method represents each column with a \emph{single} embedding vector computed from an enriched textual representation that aggregates the evidence available at ingestion time and, when enabled, an LLM-generated natural-language description.

Operationally, ColRel follows five steps: Column-to-Text construction, Column Description Generation with an LLM, Column Encoding into a single embedding per column, Similarity Computation to retrieve ranked candidate correspondences, and Human Validation of the suggested matches (and the generated descriptions).

\paragraph{Column-to-Text.}
For each column $c$, we construct a compact textual representation $colText$ from the information typically available at data-lake ingestion time. This representation combines (i) \emph{schema cues} and (ii) \emph{representative values}. Schema cues include the raw column name, tokenized variants obtained by splitting underscores and camelCase, and the declared or inferred primitive type. Representative values correspond to a fixed-size budget of examples (e.g., a mix of first non-null and frequent values), truncated to keep the input compact.

\paragraph{Column Description Generation.}
Column names and small samples are often insufficient in ERP and data-lake settings, where identifiers are opaque and conventions rely on abbreviated or coded labels. To strengthen semantic signals under these weak-metadata conditions, ColRel generates a concise natural-language description $colDesc$ for each column using an LLM.

The LLM is prompted with the column-to-text representation $colText$, augmented with enterprise knowledge when available. In particular, we exploit a business dictionary that provides code-to-label mappings, domain hints, and construction rules. If the column name (or its sub-tokens) and/or frequent values match dictionary entries, ColRel retrieves the corresponding definitions and appends them to the prompt, grounding otherwise opaque identifiers with explicit semantics. The generated description is intentionally short and strictly grounded in the provided evidence. Its goal is to verbalize the most plausible meaning of the column and make the semantics explicit for both users and the downstream matching step.

\paragraph{Column Encoding.}
We encode each column into a single dense vector using a sentence embedding model $f(\cdot)$ 
To do so, we build a final text input by concatenating the ingestion-time representation and the generated description:
\[
colInput(c) = colText(c)\ \Vert\ colDesc(c),
\]
and we set $colInput(c)=colText(c)$ when descriptions are not used or unavailable. 
The column embedding is then computed as:
\[
\mathbf{v}_c = f(colInput(c)) \in \mathbb{R}^d
\]
This design keeps the approach simple by maintaining a single vector per column while allowing the LLM-generated description to inject explicit semantic cues into the encoder input.
\paragraph{Similarity Computation.}
Given the column embeddings, ColRel scores candidate correspondences using cosine similarity. For a source column $a$ and a target column $b$, we compute:
\[
score(a,b) = \cos(\mathbf{v}_a,\mathbf{v}_b).
\]
For each source column, we rank all target columns by $score(a,b)$ and return the top-$k$ candidates as ranked suggestions. This retrieval setting matches our objective of assisting users with shortlist-based exploration rather than enforcing a strict one-to-one alignment.
\paragraph{Review and Validation.}
ColRel outputs ranked candidate correspondences for inspection. In our experiments, validation is performed by manually reviewing the top suggestions and marking them as correct or incorrect, which allows us to quantify how effectively the method surfaces true relationships under a limited review budget. 

\section{Experimental Setup}
\label{sec:experimentsetup}
Our experimental study follows a progressive evaluation designed to show the contribution of semantic enrichment while keeping the matching pipeline consistent. We compare three modeling levels:
\begin{itemize}
\item \textbf{Classical schema- and instance-based matchers}.
\item \textbf{ColRel -- Stage 1 (Embeddings only)}, where each column is encoded using ingestion-time evidence (column name, data types, and representative values), and correspondences are ranked using embedding similarity.
\item \textbf{ColRel -- Stage 2 (Semantic enrichment + embeddings)}, which extends Stage~1 by augmenting the column representation with an LLM-generated semantic description before encoding.
\end{itemize}

Stage~1 and Stage~2 share the same encoder and similarity computation pipeline. The only difference is the addition of generated descriptions in Stage~2. This controlled setup isolates the incremental effect of semantic enrichment.

Each method produces a ranked list of candidate correspondences for a given pair of tables. We evaluate performance both in a one-to-one alignment setting (for comparability with the Valentine benchmark \cite{koutras2021valentine}) and in a retrieval setting aligned with our objective: surfacing high-quality candidates for user validation.

\subsection{Datasets}
We evaluate ColRel on public benchmarks and an industrial ERP-inspired dataset. 
The public benchmarks come from Valentine~\cite{koutras2021valentine}, which is widely used in recent schema-matching evaluations~\cite{liu2024magneto,kired2024embedding,hellenberg2025prisma,kokel2025evaluating}. 
In the paper, we report results on \textbf{ChEMBL}, a biomedical dataset with specialized terminology, using both \textit{clean} and \textit{noisy} variants to assess robustness to schema noise. 
Additional WikiData results are available in the repository.

To reflect ERP conditions, we also evaluate a \textbf{Public Housing} asset-management dataset. 
Due to confidentiality constraints, we release a synthetic counterpart preserving its main schema characteristics.\footnote{Data and evaluation material: \url{https://git.msh-lse.fr/mon-groupe/colrel}.} 
It covers housing assets, identifiers, location and organizational attributes, lifecycle events, and exploitation-related information. 
Ground truth is provided by Valentine for ChEMBL and was constructed with a domain expert for Public Housing. 
The same evaluation protocol is applied across datasets.

\subsection{Evaluation Metrics}
A common approach for evaluating schema-matching methods is to assess their ability to produce high-quality ranked lists of candidate matches~\cite{koutras2021valentine,du2024situ}. This aligns with our objective of generating ranked match suggestions for users.
\textbf{Hit@5$_{\text{src}}$} checks whether at least one correct match appears within the top five suggestions, reflecting a realistic human validation budget.  
\textbf{MRR$_{\text{src}}$} (Mean Reciprocal Rank) captures ranking quality by averaging the reciprocal rank of the first correct match per source column, rewarding methods that rank true correspondences higher.
\textbf{Recall@|GT|} evaluates performance at the table-pair level by measuring how many ground-truth pairs are recovered within the top $|GT|$ globally ranked predictions.
For comparability with prior work~\cite{koutras2021valentine}, we also report \textbf{Precision}, \textbf{Recall}, and \textbf{F1} under a one-to-one alignment constraint, although this stricter regime is less aligned with our retrieval-oriented objective.

\subsection{Baselines: Classical Matchers}
We compare ColRel against classical schema matching techniques:
\textbf{Cupid}, a tree-based matcher combining linguistic and structural similarity~\cite{madhavan2001generic};
\textbf{Similarity Flooding}, a graph-based approach that propagates similarity scores to align schema elements~\cite{melnik2002similarity};
\textbf{COMA}, a composite matching framework for combining multiple matchers~\cite{do2002coma,engmann2007instance} (we use COMA 3.0 in two variants: \textbf{COMA-Schema} and \textbf{COMA-Instance});
\textbf{Distribution-Based}, an instance-based method relying on value distribution similarity~\cite{zhang2011automatic};
and \textbf{Jaccard-Levenshtein}, a value-based matcher using Jaccard similarity with approximate string equality (Levenshtein threshold) as implemented in Valentine~\cite{koutras2021valentine}.

\subsection{ColRel--Stage1: Embedding Models}
In ColRel--Stage~1, each column is encoded into a dense vector constructed from lightweight metadata (name, type, tokenized variants) and representative values. Candidate correspondences are retrieved via nearest-neighbor search in the embedding space.
We evaluate four widely used encoders available on the Hugging Face Hub:
\begin{itemize}
\item \textbf{sentence-transformers/all-MiniLM-L6-v2} (384 dimensions), a compact and efficient bi-encoder commonly used for semantic similarity and retrieval \cite{wang2020minilm}.
\item \textbf{sentence-transformers/all-MiniLM-L12-v2} (384 dimensions), a deeper MiniLM variant offering improved semantic representation while maintaining strong efficiency.
\item \textbf{sentence-transformers/all-mpnet-base-v2} (768 dimensions), a stronger general-purpose encoder that typically provides improved semantic discrimination at higher computational cost \cite{song2020mpnet}.
\item \textbf{BAAI/bge-base-en-v1.5} (768 dimensions), a retrieval-oriented model optimized for nearest-neighbor ranking quality \cite{bge_embedding}.
\end{itemize}

Model selection is guided by the MTEB benchmark \cite{muennighoff2023mteb}, which evaluates performance on retrieval and semantic similarity tasks closely related to our setting. We therefore cover a realistic spectrum of efficiency and quality: compact encoders (MiniLM), stronger general-purpose encoders (MPNet), and retrieval-focused models (BGE).

\subsection{ColRel--Stage2: Semantic Description Generation}
ColRel--Stage~2 augments the Stage~1 representation with concise natural-language descriptions generated by GPT-4o-mini (default settings, no fine-tuning). Descriptions are conditioned on available schema cues and representative values, and grounded—when available—using domain dictionaries to expand coded identifiers and constrain vocabulary. The generated description is appended to the column representation before encoding and embedded using the same encoder as in Stage~1. GPT-4o-mini was selected as a practical generation model because recent studies have used it in metadata-generation, and have shown it to be competitive with, or stronger than, open-weight alternatives in related tasks~\cite{tinn2025pre,ji2025target}.

\subsection{Computational Setup}
All experiments were run on GPU server, accessed remotely via VPN and SSH. Computations were executed on a single Nvidia RTX A6000 GPU (48\,GB VRAM).
This constraint influences our design choices in two ways:
- We focus on embedding models that can be executed efficiently on a single GPU;
- We keep column representations compact to control runtime and maintain scalability for large data lakes.

\section{Results and Discussion}
\label{sec:results}
We evaluate our approach under two relationship regimes: \textbf{joinable} and \textbf{semantically related}. In the \textbf{joinable} setting, shared identifiers and strong value overlap provide reliable instance evidence; classical matchers and embedding-based representations already perform well, so we report results for classical methods and ColRel--Stage~1 only. In the \textbf{semantically related} setting, equality joins fail because of heterogeneous conventions, coded identifiers, or limited verbatim overlap. Since lexical and instance cues are often insufficient, we first evaluate classical matchers and ColRel--Stage~1 as baselines, then activate ColRel--Stage~2 to measure the benefit of semantic enrichment through descriptions grounded in available evidence and domain dictionaries.

To keep the presentation concise and the tables readable, we omit WikiData results from the paper. WikiData is the easiest benchmark in our suite and does not meaningfully challenge the compared approaches. Full results, including WikiData and additional dataset variants, are available in the accompanying GitLab repository.
Concretely, our experiments are organized as follows: (i) \textbf{Joinable:} classical matchers vs.\ ColRel--Stage~1; (ii) \textbf{semantically related:} classical matchers vs.\ ColRel--Stage~1 vs.\ ColRel--Stage~2.

\subsection{ColRel--Stage1 Results}
Table~\ref{tab:joinable_classical_vs_stage1} reports joinable results for classical matchers and ColRel--Stage~1 (embeddings only). Since our goal is shortlist-based \emph{relationship discovery}, we focus on retrieval metrics (Rec@|GT|, Hit@5, MRR), while one-to-one (P/R/F1) is shown for comparability with Valentine~\cite{koutras2021valentine}. Overall, joinability is driven by strong instance evidence: shared identifiers and value overlap allow many methods to achieve near-perfect performance, with Hit@5 and MRR close to 1.00. Differences therefore concern ranking consistency (Rec@|GT|, MRR) and runtime rather than the mere presence of a correct match in the shortlist.

On \textbf{ChEMBL--Clean}, several classical matchers reach perfect retrieval. These include COMA-Schema, COMA-Instance, and Similarity Flooding (Rec@|GT| / Hit@5 / MRR: 1.00 / 1.00 / 1.00). ColRel--Stage~1 is similarly strong: MPNet and MiniLM-L12 achieve (Rec@|GT| / Hit@5 / MRR: 0.97 / 1.00 / 1.00). Since performance is near ceiling, runtime becomes the key differentiator: Similarity Flooding is fastest (0.70\,s), whereas Jaccard and COMA-Instance are much slower (73.58\,s and 36.57\,s). Among embeddings, MiniLM variants are the most efficient (1.01\,s and 1.96\,s), while BGE-m3 is slower (25.77\,s) despite comparable ranking quality.

On \textbf{ChEMBL--Noisy}, schema noise exposes fragility in several classical methods: COMA-Schema drops to Rec@|GT| = 0.18, and Cupid to 0.31. Jaccard becomes the strongest classical baseline on Rec@|GT| and MRR (0.78 and 0.86), confirming that value overlap remains effective in joinable settings. For embeddings, robustness is model-dependent: BGE-m3 performs best (Rec@|GT| / Hit@5 / MRR: 0.79 / 1.00 / 0.97), whereas MPNet and MiniLM-L12 show large Rec@|GT| drops (0.28 and 0.26). Their Hit@5 remains high (0.89--0.93), meaning correct matches are still retrieved but ranked less consistently at the top.

\begin{table}[!t]
\caption{Joinable results: Classical methods vs.\ ColRel--Stage~1 (embeddings only).}
\label{tab:joinable_classical_vs_stage1}
\centering
\small
\setlength{\tabcolsep}{2.5pt}
\renewcommand{\arraystretch}{1.12}

\resizebox{\linewidth}{!}{%
\begin{tabular}{l l ccc ccc c}
\toprule
\multirow{2}{*}{Dataset} & \multirow{2}{*}{Approach / Model}
& \multicolumn{3}{c}{Retrieval (primary)}
& \multicolumn{3}{c}{One-to-one (informative)}
& \multirow{2}{*}{Time (s)} \\
\cmidrule(lr){3-5}\cmidrule(lr){6-8}
& & Rec@|GT| & Hit@5 & MRR & P & R & F1 & \\
\midrule

\multirow{12}{*}{ChEMBL - Clean}
& \multicolumn{8}{l}{\textit{Classical methods}} \\
& Jaccard            & 0.78 & 0.87 & 0.87 & 0.94 & 0.78 & 0.85 & 73.58 \\
& Cupid              & 0.99 & 0.99 & 0.99 & 0.91 & 0.99 & 0.94 & 9.52 \\
& Distribution       & 0.76 & 0.71 & 0.70 & 0.91 & 0.63 & 0.73 & 1.68 \\
& COMA-Schema        & \textbf{1.00} & 1.00 & \textbf{1.00} & 0.99 & 0.75 & 0.84 & \textbf{1.21} \\
& COMA-Instance      & \textbf{1.00} & 1.00 & \textbf{1.00} & 0.99 & 0.65 & 0.77 & \textbf{36.57} \\
& Similarity Flooding & \textbf{1.00} & 1.00 & \textbf{1.00} & 0.76 & 1.00 & 0.85 & \textbf{0.70} \\
\addlinespace
& \multicolumn{8}{l}{\textit{ColRel--Stage1 (embeddings only)}} \\
& BGE-m3             & 0.93 & 1.00 & \textbf{1.00} & 0.75 & 0.99 & 0.83 & \textbf{25.77} \\
& MPNet              & \textbf{0.97} & 1.00 & \textbf{1.00} & 0.77 & 0.99 & 0.85 & \textbf{6.96} \\
& MiniLM-L12         & \textbf{0.97} & 1.00 & \textbf{1.00} & 0.75 & 0.99 & 0.84 & \textbf{1.96} \\
& MiniLM-L6          & \textbf{0.97} & 1.00 & 0.99 & 0.77 & 0.99 & 0.85 & \textbf{1.01} \\
\midrule

\multirow{12}{*}{ChEMBL - Noisy}
& \multicolumn{8}{l}{\textit{Classical methods}} \\
& Jaccard            & \textbf{0.78} & 0.86 & \textbf{0.86} & 0.93 & 0.79 & 0.85 & \textbf{69.55} \\
& Cupid              & 0.31 & 0.63 & 0.46 & 0.57 & 0.31 & 0.40 & 14.71 \\
& Distribution       & 0.71 & 0.68 & 0.67 & 0.90 & 0.62 & 0.72 & 1.58 \\
& COMA-Schema        & 0.18 & 0.12 & 0.12 & 0.63 & 0.15 & 0.24 & 1.18 \\
& COMA-Instance      & 0.69 & 0.66 & 0.66 & 0.96 & 0.47 & 0.63 & 34.01 \\
& Similarity Flooding & 0.28 & 0.96 & 0.62 & 0.39 & 0.37 & 0.38 & 0.67 \\
\addlinespace
& \multicolumn{8}{l}{\textit{ColRel--Stage1 (embeddings only)}} \\
& BGE-m3             & \textbf{0.79} & 1.00 & \textbf{0.97} & 0.72 & 0.93 & 0.80 & \textbf{25.32} \\
& MPNet              & 0.28 & 0.89 & 0.60 & 0.44 & 0.41 & 0.42 & 6.93 \\
& MiniLM-L12         & 0.26 & 0.93 & 0.74 & 0.48 & 0.53 & 0.50 & 1.94 \\
& MiniLM-L6          & 0.34 & 0.93 & 0.83 & 0.51 & 0.62 & 0.54 & 1.00 \\
\midrule

\multirow{12}{*}{Public Housing}
& \multicolumn{8}{l}{\textit{Classical methods}} \\
& Jaccard            & 0.71 & 1.00 & 1.00 & 0.76 & 0.93 & 0.84 & 2.38 \\
& Cupid              & \textbf{1.00} & 1.00 & \textbf{1.00} & 1.00 & 0.86 & 0.92 & 17.12 \\
& Distribution       & 0.71 & 1.00 & 1.00 & 0.80 & 0.57 & 0.67 & 2.76 \\
& COMA-Schema        & \textbf{1.00} & 1.00 & \textbf{1.00} & 1.00 & 1.00 & 1.00 & \textbf{1.47} \\
& COMA-Instance      & \textbf{1.00} & 1.00 & \textbf{1.00} & 1.00 & 0.79 & 0.88 & 10.44 \\
& Similarity Flooding & \textbf{1.00} & 1.00 & \textbf{1.00} & 0.70 & 1.00 & 0.82 & 4.07 \\
\addlinespace
& \multicolumn{8}{l}{\textit{ColRel--Stage1 (embeddings only)}} \\
& BGE-m3             & \textbf{0.86} & 1.00 & \textbf{1.00} & 0.64 & 1.00 & 0.78 & 47.73 \\
& MPNet              & 0.43 & 1.00 & 0.96 & 0.65 & 0.93 & 0.76 & 13.82 \\
& MiniLM-L12         & 0.79 & 1.00 & \textbf{1.00} & 0.67 & 1.00 & 0.80 & 13.21 \\
& MiniLM-L6          & 0.71 & 1.00 & \textbf{1.00} & 0.70 & 1.00 & 0.82 & \textbf{4.16} \\
\bottomrule
\end{tabular}%
}

\end{table}
\FloatBarrier

On \textbf{Public Housing}, although schema naming remains difficult, the joinable regime is still relatively easy because value overlap remains strong. Several classical methods achieve perfect retrieval, notably Cupid, COMA-Schema, COMA-Instance, and Similarity Flooding, while Jaccard and Distribution-Based remain slightly below ceiling on Rec@|GT| (0.71). ColRel--Stage~1 also produces strong shortlists, with Hit@5 = 1.00 for all encoders and MRR reaching 1.00 for BGE-m3, MiniLM-L12, and MiniLM-L6. However, embedding models remain below the strongest classical methods on Rec@|GT|, with BGE-m3 performing best on this metric (0.86), followed by MiniLM-L12 (0.79), MiniLM-L6 (0.71), and MPNet (0.43). 
In summary, joinable cases are largely driven by instance overlap; differences mainly concern robustness to schema noise, ranking consistency, and efficiency. This motivates studying semantically related cases, where semantic enrichment (ColRel--Stage~2) is most beneficial.

\subsection{ColRel--Stage2 Results}
\label{sec:semjoin_stage2_results}
Table~\ref{tab:semjoin_classical_stage1_stage2} reports results in the \emph{semantically related} regime, which is more challenging than joinable cases because equality joins fail under heterogeneous conventions, coded identifiers, and weak lexical overlap. In this setting, instance- and string-based signals are often insufficient, and ranking consistency becomes the main difficulty.

On \textbf{ChEMBL--Clean}, some classical methods still achieve perfect retrieval, including Cupid, Similarity Flooding, and COMA-Instance (Rec@|GT| / Hit@5 / MRR: 1.00 / 1.00 / 1.00), showing that clean schemata retain exploitable structure. ColRel--Stage~1 already performs strongly, with MiniLM-L12 and BGE-m3 reaching Rec@|GT| = 0.84 and MRR up to 0.98. ColRel--Stage~2 further improves ranking stability: MiniLM-L12 and MPNet reach Rec@|GT| = 0.90, and MiniLM-L12 achieves MRR = 1.00. Although gains remain moderate because the baseline is already high, they confirm that semantic enrichment improves ranking quality without major runtime overhead (e.g., 1.43\,s and 0.72\,s for MiniLM-L12 and MiniLM-L6).

The difference becomes clearer on \textbf{ChEMBL--Noisy}. Classical methods degrade sharply; the best baseline is Jaccard (Rec@|GT| / MRR: 0.57 / 0.81). ColRel--Stage~1 already mitigates this degradation, with BGE-m3 achieving the strongest retrieval scores overall (Rec@|GT| / Hit@5 / MRR: 0.66 / 0.97 / 0.94). ColRel--Stage~2 further improves several encoder variants: MiniLM-L6, MiniLM-L12, and MPNet gain in Rec@|GT| while maintaining strong shortlist quality, although gains are not uniform across models. For noisy ChEMBL, many column names are coded. To support description generation, we manually constructed a controlled dictionary from the public ChEMBL schema documentation, without expert assistance, and used it to expand coded identifiers before prompting. Although this dictionary was not validated by domain experts, the results suggest that even documentation-based lexical grounding can improve robustness in coded schemata. This strengthens our motivation for ERP settings, where expert-validated business dictionaries are likely to provide even more reliable support for semantic enrichment.

\begin{table}[!t]
\caption{Semantically related results: Classical methods vs.\ ColRel--Stage~1 vs.\ ColRel--Stage~2.}
\label{tab:semjoin_classical_stage1_stage2}
\centering
\small
\setlength{\tabcolsep}{3pt}
\renewcommand{\arraystretch}{1.12}

\resizebox{\linewidth}{!}{%
\begin{tabular}{l l ccc ccc c}
\toprule
\multirow{2}{*}{Dataset} & \multirow{2}{*}{Approach / Model}
& \multicolumn{3}{c}{Retrieval (primary)}
& \multicolumn{3}{c}{One-to-one (informative)}
& \multirow{2}{*}{Time (s)} \\
\cmidrule(lr){3-5}\cmidrule(lr){6-8}
& & Rec@|GT| & Hit@5 & MRR & P & R & F1 & \\
\midrule

\multirow{1}{*}{ChEMBL -- Clean} 
& \multicolumn{8}{l}{\textit{Classical methods}} \\

& Cupid 
& \textbf{1.00} & \textbf{1.00} & \textbf{1.00}
& 0.89 & 1.00 & 0.93 & \textbf{10.33} \\

& Similarity Flooding 
& \textbf{1.00} & \textbf{1.00} & \textbf{1.00}
& 0.73 & 1.00 & 0.83 & \textbf{0.75} \\

& COMA-Instance 
& \textbf{1.00} & \textbf{1.00} & \textbf{1.00}
& 0.99 & 0.57 & 0.71 & \textbf{39.61} \\

& Distribution-Based 
& 0.35 & 0.37 & 0.36
& 0.83 & 0.29 & 0.42 & 2.46 \\

& Jaccard 
& 0.54 & 0.80 & 0.76
& 0.77 & 0.50 & 0.60 & 96.66 \\

\addlinespace
& \multicolumn{8}{l}{\textit{ColRel--Stage1}} \\

& MiniLM-L6 
& 0.81 & 0.97 & 0.97
& 0.71 & 0.93 & 0.79 & 1.19 \\

& MiniLM-L12 
& \textbf{0.84} & 0.97 & 0.97
& 0.74 & 0.94 & 0.81 & \textbf{2.29} \\

& MPNet 
& 0.79 & 0.97 & 0.97
& 0.68 & 0.90 & 0.76 & 8.20 \\

& BGE-m3 
& \textbf{0.84} & \textbf{0.98} & \textbf{0.98}
& 0.73 & 0.96 & 0.81 & \textbf{28.61} \\

\addlinespace
& \multicolumn{8}{l}{\textit{ColRel--Stage2}} \\

& MiniLM-L6 
& 0.87 & 0.98 & 0.99
& 0.76 & 0.97 & 0.84 & 0.72 \\

& MiniLM-L12 
& \textbf{0.90} & \textbf{0.99} & \textbf{1.00}
& 0.74 & 0.96 & 0.82 & \textbf{1.43} \\

& MPNet 
& \textbf{0.90} & 0.96 & 0.98
& 0.75 & 0.97 & 0.83 & 5.01 \\

& BGE-m3 
& 0.84 & 0.97 & 0.99
& 0.74 & 0.94 & 0.82 & 21.90 \\

\midrule
\multirow{1}{*}{ChEMBL -- Noisy} 
& \multicolumn{8}{l}{\textit{Classical methods}} \\

& Jaccard 
& \textbf{0.57} & \textbf{0.81} & \textbf{0.81}
& 0.78 & 0.53 & 0.62 & \textbf{78.08} \\

& Distribution-Based 
& 0.43 & 0.47 & 0.45
& 0.82 & 0.35 & 0.47 & 2.04 \\

& COMA-Instance 
& 0.49 & 0.51 & 0.51
& 0.86 & 0.34 & 0.47 & 38.41 \\

& Cupid 
& 0.28 & 0.59 & 0.43
& 0.65 & 0.29 & 0.40 & 14.59 \\

& Similarity Flooding 
& 0.19 & \textbf{0.96} & 0.62
& 0.18 & 0.16 & 0.17 & 0.67 \\

\addlinespace
& \multicolumn{8}{l}{\textit{ColRel--Stage1}} \\

& MiniLM-L6 
& 0.34 & 0.85 & 0.79
& 0.45 & 0.51 & 0.47 & 1.15 \\

& MiniLM-L12 
& 0.25 & 0.84 & 0.65
& 0.34 & 0.38 & 0.35 & 2.16 \\

& MPNet 
& 0.19 & 0.80 & 0.57
& 0.31 & 0.29 & 0.30 & 7.86 \\

& BGE-m3 
& \textbf{0.66} & \textbf{0.97} & \textbf{0.94}
& 0.65 & 0.82 & 0.72 & \textbf{27.71} \\

\addlinespace
& \multicolumn{8}{l}{\textit{ColRel--Stage2}} \\

& MiniLM-L6 
& 0.41 & \textbf{0.96} & 0.79
& 0.49 & 0.65 & 0.55 & 0.99 \\

& MiniLM-L12 
& 0.47 & 0.94 & 0.73
& 0.48 & 0.63 & 0.54 & 2.45 \\

& MPNet 
& 0.54 & 0.94 & 0.71
& 0.51 & 0.66 & 0.57 & 5.68 \\

& BGE-m3 
& \textbf{0.59} & 0.95 & \textbf{0.81}
& 0.55 & 0.74 & 0.62 & \textbf{23.79} \\

\midrule
\multirow{1}{*}{Public Housing} 
& \multicolumn{8}{l}{\textit{Classical methods}} \\

& COMA-Schema 
& \textbf{0.41} & 0.41 & 0.41
& 0.44 & 0.41 & 0.42 & \textbf{2.95} \\

& COMA-Instance 
& 0.24 & 0.24 & 0.24
& 0.36 & 0.24 & 0.29 & 115.24 \\

& Similarity Flooding 
& 0.29 & 0.53 & 0.53
& 0.16 & 0.41 & 0.23 & 36.47 \\

& Distribution-Based 
& 0.06 & 0.12 & 0.09
& 0.12 & 0.06 & 0.08 & 19.38 \\

& Jaccard 
& 0.00 & 0.12 & 0.06
& 0.07 & 0.06 & 0.06 & 18.99 \\

& Cupid 
& 0.00 & 0.00 & 0.00
& 0.00 & 0.00 & 0.00 & 422.94 \\

\addlinespace
& \multicolumn{8}{l}{\textit{ColRel--Stage1}} \\

& MiniLM-L6 
& 0.00 & 0.65 & 0.41
& 0.00 & 0.00 & 0.00 & \textbf{5.90} \\

& MiniLM-L12 
& 0.00 & 0.53 & 0.43
& 0.07 & 0.18 & 0.10 & 11.22 \\

& MPNet 
& 0.06 & 0.41 & 0.36
& 0.07 & 0.18 & 0.10 & 39.86 \\

& BGE-m3 
& 0.06 & 0.65 & 0.37
& 0.04 & 0.12 & 0.06 & 104.45 \\

\addlinespace
& \multicolumn{8}{l}{\textit{ColRel--Stage2}} \\

& BGE-m3 
& \textbf{0.35} & \textbf{0.82} & \textbf{0.68}
& 0.16 & 0.41 & 0.23 & 154.31 \\

& MPNet 
& 0.12 & 0.71 & 0.57
& 0.07 & 0.18 & 0.10 & 51.46 \\

& MiniLM-L12 
& 0.12 & 0.71 & 0.56
& 0.09 & 0.24 & 0.13 & 15.31 \\

& MiniLM-L6 
& 0.24 & 0.71 & 0.51
& 0.11 & 0.29 & 0.16 & 8.58 \\

\bottomrule
\end{tabular}%
}

\end{table}
\FloatBarrier

The clearest effect appears on \textbf{Public Housing}. In this setting, the key objective is to maximize useful matches within the shortlist presented to the user, rather than to enforce a fully automatic one-to-one alignment. From this perspective, classical methods remain limited: the best classical shortlist quality is obtained by Similarity Flooding (Hit@5 / MRR: 0.53 / 0.53), while COMA-Schema reaches only (Hit@5 / MRR: 0.41 / 0.41). ColRel--Stage~1 already improves the shortlist, with Hit@5 up to 0.65 and MRR up to 0.43, but rankings remain unstable. ColRel--Stage~2 makes the effect much clearer: BGE-m3 reaches (Hit@5 / MRR: 0.82 / 0.68), meaning that relevant target columns are much more often retrieved within the top candidates and ranked earlier in the list. The other Stage2 variants also remain consistently stronger than Stage1 on shortlist usefulness, with Hit@5 = 0.71 and MRR between 0.51 and 0.57. In coded ERP schemata, semantic enrichment is therefore essential to maximize useful matches in the shortlist and make user-driven validation much more effective.
\section{Conclusion and Future Work}
\label{sec:conclusion}
This paper addresses column-level relationship discovery in data lakes under industrial constraints, with a focus on ERP-derived data characterized by coded schemata and weak metadata. We introduced ColRel, a two-stage method that performs embedding-based retrieval and optionally enriches column representations with business dictionary and LLM-generated descriptions.

Our experiments show that joinable settings are often close to ceiling due to strong value overlap, whereas semantically related settings remain challenging when lexical overlap is weak and equality joins fail. In this regime, ColRel--Stage2 provides clear gains: grounded semantic enrichment improves ranking quality and shortlist coherence without changing the embedding and similarity computation pipeline.

Although our evaluation focuses on ERP data, ColRel can be applied to other fields whenever columns can be represented through a combination of schema cues, representative values, and business dictionary. In practice, this means that the method remains generic, while its adaptation to a new domain mainly requires an appropriate user- or expert-provided dictionary, terminology resource, or equivalent grounding support.

Several directions remain open for future work. First, we plan to implement ColRel into an interactive assistant for data lake exploration, where user feedback on suggested correspondences can be captured and reused to refine rankings. Second, description generation could be improved through better prompt design, ontology-based guidance, or other grounding mechanisms, in order to reduce reliance on manual dictionaries while preserving robustness. Third, we intend to investigate scalability aspects, including indexing strategies and incremental updates for very large data lakes. Fourth, privacy and confidentiality issues also deserve further investigation, since description generation may require limited schema cues and a small number of representative values when metadata alone are insufficient. In this respect, locally deployable language models could provide a useful alternative to reduce exposure to external services while preserving domain-specific semantic enrichment. Finally, beyond relationship discovery, generated semantic descriptions could support broader metadata enrichment goals, including data cataloging and governance within industrial data lakes~\cite{deng2024lakebench}.



\begin{credits}
\subsubsection{\ackname} 
Ahlame Diouan’s PhD is funded by BIAL-X.

\subsubsection{\discintname}
The authors have no competing interests to declare that are relevant to the content of this article.
\end{credits}

%
%

\nocite{*}
\bibliographystyle{splncs04}
\bibliography{biblio}

\begin{thebibliography}{10}
\providecommand{\url}[1]{\texttt{#1}}
\providecommand{\urlprefix}{URL }
\providecommand{\doi}[1]{https://doi.org/#1}

\bibitem{bogatu2020dataset}
Bogatu, A., Fernandes, A.A., Paton, N.W., Konstantinou, N.: Dataset discovery
  in data lakes. In: 2020 ieee 36th international conference on data
  engineering (icde). pp. 709--720. IEEE (2020)

\bibitem{cappuzzo2021embdi}
Cappuzzo, R., Papotti, P., Thirumuruganathan, S.: Embdi: generating embeddings
  for relational data integration. In: 29th Italian Symposium on Advanced
  Database Systems (SEDB), Pizzo Calabro, Italy (2021)

\bibitem{deng2024lakebench}
Deng, Y., Chai, C., Cao, L., Yuan, Q., Chen, S., Yu, Y., Sun, Z., Wang, J., Li,
  J., Cao, Z., et~al.: Lakebench: A benchmark for discovering joinable and
  unionable tables in data lakes. Proceedings of the VLDB Endowment
  \textbf{17}(8),  1925--1938 (2024)

\bibitem{diouan2024relationships}
Diouan, A., Ferey, E., Darmont, J., Loudcher, S.: About relationships in data
  lakes. In: International Database Engineered Applications Symposium. pp.
  141--155. Springer (2024)

\bibitem{diouan2022metadonnees}
Diouan, A., Ferey, E., Loudcher, S., Darmont, J., No{\^u}s, C.:
  M{\'e}tadonn{\'e}es des lacs de donn{\'e}es et principes fair. In: EDA. pp.
  109--110 (2022)

\bibitem{do2002coma}
Do, H.H., Rahm, E.: Coma—a system for flexible combination of schema matching
  approaches. In: VLDB'02: Proceedings of the 28th International Conference on
  Very Large Databases. pp. 610--621. Elsevier (2002)

\bibitem{doan2005semantic}
Doan, A., Halevy, A.Y.: Semantic integration research in the database
  community: A brief survey. AI magazine  \textbf{26}(1),  83--83 (2005)

\bibitem{dong2022deepjoin}
Dong, Y., Xiao, C., Nozawa, T., Enomoto, M., Oyamada, M.: Deepjoin: Joinable
  table discovery with pre-trained language models. arXiv preprint
  arXiv:2212.07588  (2022)

\bibitem{du2024situ}
Du, X., Yuan, G., Wu, S., Chen, G., Lu, P.: In situ neural relational schema
  matcher. In: 2024 IEEE 40th International Conference on Data Engineering
  (ICDE). pp. 138--150. IEEE (2024)

\bibitem{engmann2007instance}
Engmann, D., Massmann, S.: Instance matching with coma++. In: BTW workshops.
  vol.~7, pp. 28--37 (2007)

\bibitem{fernandez2018aurum}
Fernandez, R.C., Abedjan, Z., Koko, F., Yuan, G., Madden, S., Stonebraker, M.:
  Aurum: A data discovery system. In: 2018 IEEE 34th International conference
  on data engineering (ICDE). pp. 1001--1012. IEEE (2018)

\bibitem{hellenberg2025prisma}
Hellenberg, J.E., Mahling, F.D., Laskowski, L., Naumann, F., Paganelli, M.,
  Panse, F., et~al.: Prisma: A privacy-preserving schema matcher using
  functional dependencies. In: Advances in Database Technology-EDBT, vol.~28,
  pp. 297--309. OpenProceedings. org (2025)

\bibitem{ji2025target}
Ji, X., Glenn, P., Parameswaran, A.G., Hulsebos, M.: Target: Benchmarking table
  retrieval for generative tasks. arXiv preprint arXiv:2505.11545  (2025)

\bibitem{jiang2023mistral7b}
Jiang, A.Q., Sablayrolles, A., Mensch, A., et~al.: Mistral 7b (2023)

\bibitem{kired2024embedding}
Kired, N.E., Ravat, F., Song, J., Teste, O.: Embedding-based data matching for
  disparate data sources. In: International Conference on Big Data Analytics
  and Knowledge Discovery. pp. 66--71. Springer (2024)

\bibitem{kokel2025evaluating}
Kokel, H., Khatiwada, A., Pedapati, T., Ananthakrishnan, H., Hassanzadeh, O.,
  Samulowitz, H., Srinivas, K.: Evaluating joinable column discovery approaches
  for context-aware search. arXiv preprint arXiv:2510.24599  (2025)

\bibitem{koutras2021valentine}
Koutras, C., Siachamis, G., Ionescu, A., Psarakis, K., Brons, J., Fragkoulis,
  M., Lofi, C., Bonifati, A., Katsifodimos, A.: Valentine: Evaluating matching
  techniques for dataset discovery. In: 2021 IEEE 37th International Conference
  on Data Engineering (ICDE). pp. 468--479. IEEE (2021)

\bibitem{liu2024magneto}
Liu, Y., Pena, E., Santos, A., Wu, E., Freire, J.: Magneto: Combining small and
  large language models for schema matching. arXiv preprint arXiv:2412.08194
  (2024)

\bibitem{madhavan2001generic}
Madhavan, J., Bernstein, P.A., Rahm, E., et~al.: Generic schema matching with
  cupid. In: vldb. vol.~1, pp. 49--58 (2001)

\bibitem{melnik2002similarity}
Melnik, S., Garcia-Molina, H., Rahm, E.: Similarity flooding: A versatile graph
  matching algorithm and its application to schema matching. In: Proceedings
  18th international conference on data engineering. pp. 117--128. IEEE (2002)

\bibitem{muennighoff2023mteb}
Muennighoff, N., Tazi, N., Magne, L., Reimers, N.: Mteb: Massive text embedding
  benchmark. In: Proceedings of the 17th Conference of the European Chapter of
  the Association for Computational Linguistics. pp. 2014--2037 (2023)

\bibitem{parciak2024schema}
Parciak, M., Vandevoort, B., Neven, F., Peeters, L.M., Vansummeren, S.: Schema
  matching with large language models: an experimental study. arXiv preprint
  arXiv:2407.11852  (2024)

\bibitem{rahm2001survey}
Rahm, E., Bernstein, P.A.: A survey of approaches to automatic schema matching.
  the VLDB Journal  \textbf{10}(4),  334--350 (2001)

\bibitem{reimers2019sentence}
Reimers, N., Gurevych, I.: Sentence-bert: Sentence embeddings using siamese
  bert-networks. arXiv preprint arXiv:1908.10084  (2019)

\bibitem{sheetrit2024rematch}
Sheetrit, E., Brief, M., Mishaeli, M., Elisha, O.: Rematch: Retrieval enhanced
  schema matching with llms. arXiv preprint arXiv:2403.01567  (2024)

\bibitem{song2020mpnet}
Song, K., Tan, X., Qin, T., Lu, J., Liu, T.Y.: Mpnet: Masked and permuted
  pre-training for language understanding. Advances in neural information
  processing systems  \textbf{33},  16857--16867 (2020)

\bibitem{tinn2025pre}
Tinn, P., S{\o}rb{\o}, S., Jiang, S., Voutetakis, K., Giounis, S.M., Pilalis,
  E., Papadodima, O., Roman, D.: Pre-meta: priors-augmented retrieval for
  llm-based metadata generation. Bioinformatics  \textbf{41}(10),  btaf519
  (2025)

\bibitem{wang2020minilm}
Wang, W., Wei, F., Dong, L., Bao, H., Yang, N., Zhou, M.: Minilm: Deep
  self-attention distillation for task-agnostic compression of pre-trained
  transformers. Advances in neural information processing systems  \textbf{33},
   5776--5788 (2020)

\bibitem{bge_embedding}
Xiao, S., Liu, Z., Zhang, P., Muennighoff, N.: C-pack: Packaged resources to
  advance general chinese embedding (2023)

\bibitem{zhang2011automatic}
Zhang, M., Hadjieleftheriou, M., Ooi, B.C., Procopiuc, C.M., Srivastava, D.:
  Automatic discovery of attributes in relational databases. In: Proceedings of
  the 2011 ACM SIGMOD International Conference on Management of data. pp.
  109--120 (2011)

\bibitem{zhang2025smutf}
Zhang, Y., Mei, D., Luo, H., Xu, C., Tsai, R.T.H.: Smutf: Schema matching using
  generative tags and hybrid features. Information Systems  \textbf{133},
  102570 (2025)

\end{thebibliography}


\bibitem[Kou+21]{koutras2021valentine}
{\au{C. Koutras}, \au{G. Siachamis}, \au{A. Ionescu}, \au{K. Psarakis}, \au{J. Brons}, \au{M. Fragkoulis}, \au{C. Lofi}, \au{A. Bonifati}, and \au{A. Katsifodimos}}, \at{\textit{Valentine: Evaluating matching techniques for dataset discovery}}, \bt{2021 IEEE 37th International Conference on Data Engineering (ICDE)}, (\yr{2021}), \pg{{468}--{479}}.

\bibitem[DR02]{do2002coma}
{\au{H. H. Do} and \au{E. Rahm}}, \at{\textit{COMA—a system for flexible combination of schema matching approaches}}, \bt{VLDB'02: Proceedings of the 28th International Conference on Very Large Databases}, (\yr{2002}), \pg{{610}--{621}}.

\bibitem[EM07]{engmann2007instance}
{\au{D. Engmann} and \au{S. Massmann}}, \at{\textit{Instance Matching with COMA++.}}, \bt{BTW workshops}, \bvol{7}, (\yr{2007}), \pg{{28}--{37}}.

\bibitem[Mad+01]{madhavan2001generic}
{\au{J. Madhavan}, \au{P. A. Bernstein}, \au{E. Rahm}, and \au{others}}, \at{\textit{Generic schema matching with cupid}}, \bt{vldb}, \bvol{1}, (\yr{2001}), \pg{{49}--{58}}.

\bibitem[MGR02]{melnik2002similarity}
{\au{S. Melnik}, \au{H. Garcia-Molina}, and \au{E. Rahm}}, \at{\textit{Similarity flooding: A versatile graph matching algorithm and its application to schema matching}}, \bt{Proceedings 18th international conference on data engineering}, (\yr{2002}), \pg{{117}--{128}}.

\bibitem[Zha+11]{zhang2011automatic}
{\au{M. Zhang}, \au{M. Hadjieleftheriou}, \au{B. C. Ooi}, \au{C. M. Procopiuc}, and \au{D. Srivastava}}, \at{\textit{Automatic discovery of attributes in relational databases}}, \bt{Proceedings of the 2011 ACM SIGMOD International Conference on Management of data}, (\yr{2011}), \pg{{109}--{120}}.

\bibitem[RG19]{reimers2019sentence}
{\au{N. Reimers} and \au{I. Gurevych}}, \at{\textit{Sentence-bert: Sentence embeddings using siamese bert-networks}}, \jt{arXiv preprint arXiv:1908.10084} (\yr{2019}).

\bibitem[RB01]{rahm2001survey}
{\au{E. Rahm} and \au{P. A. Bernstein}}, \at{\textit{A survey of approaches to automatic schema matching}}, \jt{the VLDB Journal}, \bvol{10}(4) (\yr{2001}) \pg{{334}--{350}}.

\bibitem[DH05]{doan2005semantic}
{\au{A. Doan} and \au{A. Y. Halevy}}, \at{\textit{Semantic integration research in the database community: A brief survey}}, \jt{AI magazine}, \bvol{26}(1) (\yr{2005}) \pg{{83}--{83}}.

\bibitem[Fer+18]{fernandez2018aurum}
{\au{R. C. Fernandez}, \au{Z. Abedjan}, \au{F. Koko}, \au{G. Yuan}, \au{S. Madden}, and \au{M. Stonebraker}}, \at{\textit{Aurum: A data discovery system}}, \bt{2018 IEEE 34th International conference on data engineering (ICDE)}, (\yr{2018}), \pg{{1001}--{1012}}.

\bibitem[Bog+20]{bogatu2020dataset}
{\au{A. Bogatu}, \au{A. A. Fernandes}, \au{N. W. Paton}, and \au{N. Konstantinou}}, \at{\textit{Dataset discovery in data lakes}}, \bt{2020 ieee 36th international conference on data engineering (icde)}, (\yr{2020}), \pg{{709}--{720}}.

\bibitem[Den+24]{deng2024lakebench}
{\au{Y. Deng}, \au{C. Chai}, \au{L. Cao}, \au{Q. Yuan}, \au{S. Chen}, \au{Y. Yu}, \au{Z. Sun}, \au{J. Wang}, \au{J. Li}, \au{Z. Cao}, and \au{others}}, \at{\textit{Lakebench: A benchmark for discovering joinable and unionable tables in data lakes}}, \jt{Proceedings of the VLDB Endowment}, \bvol{17}(8) (\yr{2024}) \pg{{1925}--{1938}}.

\bibitem[CPT21]{cappuzzo2021embdi}
{\au{R. Cappuzzo}, \au{P. Papotti}, and \au{S. Thirumuruganathan}}, \at{\textit{Embdi: generating embeddings for relational data integration}}, \bt{29th Italian Symposium on Advanced Database Systems (SEDB), Pizzo Calabro, Italy}, (\yr{2021}).

\bibitem[Don+22]{dong2022deepjoin}
{\au{Y. Dong}, \au{C. Xiao}, \au{T. Nozawa}, \au{M. Enomoto}, and \au{M. Oyamada}}, \at{\textit{Deepjoin: Joinable table discovery with pre-trained language models}}, \jt{arXiv preprint arXiv:2212.07588} (\yr{2022}).

\bibitem[Par+24]{parciak2024schema}
{\au{M. Parciak}, \au{B. Vandevoort}, \au{F. Neven}, \au{L. M. Peeters}, and \au{S. Vansummeren}}, \at{\textit{Schema matching with large language models: an experimental study}}, \jt{arXiv preprint arXiv:2407.11852} (\yr{2024}).

\bibitem[She+24]{sheetrit2024rematch}
{\au{E. Sheetrit}, \au{M. Brief}, \au{M. Mishaeli}, and \au{O. Elisha}}, \at{\textit{Rematch: Retrieval enhanced schema matching with llms}}, \jt{arXiv preprint arXiv:2403.01567} (\yr{2024}).

\bibitem[Liu+24]{liu2024magneto}
{\au{Y. Liu}, \au{E. Pena}, \au{A. Santos}, \au{E. Wu}, and \au{J. Freire}}, \at{\textit{Magneto: Combining small and large language models for schema matching}}, \jt{arXiv preprint arXiv:2412.08194} (\yr{2024}).

\bibitem[Zha+25]{zhang2025smutf}
{\au{Y. Zhang}, \au{D. Mei}, \au{H. Luo}, \au{C. Xu}, and \au{R. T. H. Tsai}}, \at{\textit{Smutf: Schema matching using generative tags and hybrid features}}, \jt{Information Systems}, \bvol{133} (\yr{2025}) \pg{102570}.

\bibitem[Mue+23]{muennighoff2023mteb}
{\au{N. Muennighoff}, \au{N. Tazi}, \au{L. Magne}, and \au{N. Reimers}}, \at{\textit{Mteb: Massive text embedding benchmark}}, \bt{Proceedings of the 17th Conference of the European Chapter of the Association for Computational Linguistics}, (\yr{2023}), \pg{{2014}--{2037}}.

\bibitem[Son+20]{song2020mpnet}
{\au{K. Song}, \au{X. Tan}, \au{T. Qin}, \au{J. Lu}, and \au{T. Y. Liu}}, \at{\textit{Mpnet: Masked and permuted pre-training for language understanding}}, \jt{Advances in neural information processing systems}, \bvol{33} (\yr{2020}) \pg{{16857}--{16867}}.

\bibitem[Wan+20]{wang2020minilm}
{\au{W. Wang}, \au{F. Wei}, \au{L. Dong}, \au{H. Bao}, \au{N. Yang}, and \au{M. Zhou}}, \at{\textit{Minilm: Deep self-attention distillation for task-agnostic compression of pre-trained transformers}}, \jt{Advances in neural information processing systems}, \bvol{33} (\yr{2020}) \pg{{5776}--{5788}}.

\bibitem[Xia+23]{bge_embedding}
{\au{S. Xiao}, \au{Z. Liu}, \au{P. Zhang}, and \au{N. Muennighoff}}, \at{\textit{C-Pack: Packaged Resources To Advance General Chinese Embedding}}, preprint, (\yr{2023}). arXiv:\arxiv{2309.07597}.

\bibitem[Jia+23]{jiang2023mistral7b}
{\au{A. Q. Jiang}, \au{A. Sablayrolles}, \au{A. Mensch}, and \au{others}}, \at{\textit{Mistral 7B}}, preprint, (\yr{2023}). arXiv:\arxiv{2310.06825}.

\bibitem[Dio+24]{diouan2024relationships}
{\au{A. Diouan}, \au{E. Ferey}, \au{J. Darmont}, and \au{S. Loudcher}}, \at{\textit{About Relationships in Data Lakes}}, \bt{International Database Engineered Applications Symposium}, (\yr{2024}), \pg{{141}--{155}}.

\bibitem[Kir+24]{kired2024embedding}
{\au{N. E. Kired}, \au{F. Ravat}, \au{J. Song}, and \au{O. Teste}}, \at{\textit{Embedding-based data matching for disparate data sources}}, \bt{International Conference on Big Data Analytics and Knowledge Discovery}, (\yr{2024}), \pg{{66}--{71}}.

\bibitem[Hel+25]{hellenberg2025prisma}
{\au{J. E. Hellenberg}, \au{F. D. Mahling}, \au{L. Laskowski}, \au{F. Naumann}, \au{M. Paganelli}, \au{F. Panse}, and \au{others}}, \at{\textit{Prisma: A privacy-preserving schema matcher using functional dependencies}}, \bt{Advances in Database Technology-EDBT}, \bvol{28}, (\yr{2025}), \pg{{297}--{309}}.

\bibitem[Kok+25]{kokel2025evaluating}
{\au{H. Kokel}, \au{A. Khatiwada}, \au{T. Pedapati}, \au{H. Ananthakrishnan}, \au{O. Hassanzadeh}, \au{H. Samulowitz}, and \au{K. Srinivas}}, \at{\textit{Evaluating Joinable Column Discovery Approaches for Context-Aware Search}}, \jt{arXiv preprint arXiv:2510.24599} (\yr{2025}).

\bibitem[Dio+22]{diouan2022metadonnees}
{\au{A. Diouan}, \au{E. Ferey}, \au{S. Loudcher}, \au{J. Darmont}, and \au{C. No\^us}}, \at{\textit{M\'etadonn\'ees des lacs de donn\'ees et principes FAIR.}}, \bt{EDA}, (\yr{2022}), \pg{{109}--{110}}.

\bibitem[Du+24]{du2024situ}
{\au{X. Du}, \au{G. Yuan}, \au{S. Wu}, \au{G. Chen}, and \au{P. Lu}}, \at{\textit{In situ neural relational schema matcher}}, \bt{2024 IEEE 40th International Conference on Data Engineering (ICDE)}, (\yr{2024}), \pg{{138}--{150}}.

\bibitem[Ji+25]{ji2025target}
{\au{X. Ji}, \au{P. Glenn}, \au{A. G. Parameswaran}, and \au{M. Hulsebos}}, \at{\textit{TARGET: Benchmarking table retrieval for generative tasks}}, \jt{arXiv preprint arXiv:2505.11545} (\yr{2025}).

\bibitem[Tin+25]{tinn2025pre}
{\au{P. Tinn}, \au{S. S\orb\o}, \au{S. Jiang}, \au{K. Voutetakis}, \au{S. M. Giounis}, \au{E. Pilalis}, \au{O. Papadodima}, and \au{D. Roman}}, \at{\textit{Pre-Meta: priors-augmented retrieval for LLM-based metadata generation}}, \jt{Bioinformatics}, \bvol{41}(10) (\yr{2025}) \pg{btaf519}.

\end{document}